\documentclass{article} 
\usepackage{iclr2024_conference,times}

\usepackage{amsmath,amsfonts,bm}

\def\eqref#1{equation~\ref{#1}}

\def\1{\bm{1}}

\DeclareMathAlphabet{\mathsfit}{\encodingdefault}{\sfdefault}{m}{sl}
\SetMathAlphabet{\mathsfit}{bold}{\encodingdefault}{\sfdefault}{bx}{n}

\usepackage{hyperref}
\usepackage{url}
\usepackage{times}
\usepackage{epsfig}
\usepackage{graphicx}
\usepackage{amsmath}
\usepackage{amssymb}
\usepackage{subfigure}
\usepackage{multirow}
\usepackage{wrapfig}
\usepackage{mathabx}

\usepackage{microtype}
\usepackage{graphicx}
\usepackage{booktabs} 
\usepackage{bbm,multirow,listings}
\usepackage{enumitem}
\usepackage{appendix}
\usepackage{todonotes}

\usepackage{algorithm,algpseudocode}
\algnewcommand\algorithmicinput{\textbf{Input:}}
\algnewcommand\INPUT{\item[\algorithmicinput]}
\algnewcommand\algorithmicinit{\textbf{Initialize:}}
\algnewcommand\Initialize{\item[\algorithmicinit]}
\algrenewcommand\algorithmicreturn{\textbf{Return:}}
\algnewcommand\RETURN{\item[\algorithmicreturn]}

\title{A Variational Perspective on Solving Inverse Problems with Diffusion Models}

\author{Morteza Mardani, Jiaming Song, Jan Kautz, Arash Vahdat\\
NVIDIA Inc.\\
{\tt mmardani,jiamings,jkautz,avahdat@nvidia.com}
}

\iclrfinalcopy 

\begin{document}

\maketitle
\vspace{-3mm}
\begin{abstract}
Diffusion models have emerged as a key pillar of foundation models in visual domains. One of their critical applications is to universally solve different downstream inverse tasks via a single diffusion prior without re-training for each task. Most inverse tasks can be formulated as inferring a posterior distribution over data (e.g., a full image) given a measurement (e.g., a masked image). This is however challenging in diffusion models since the nonlinear and iterative nature of the diffusion process renders the posterior intractable. To cope with this challenge, we propose a variational approach that by design seeks to approximate the true posterior distribution. We show that our approach naturally leads to regularization by denoising diffusion process (RED-diff) where denoisers at different timesteps concurrently impose different structural constraints over the image. To gauge the contribution of denoisers from different timesteps, we propose a weighting mechanism based on signal-to-noise-ratio (SNR). Our approach provides a new variational perspective for solving inverse problems with diffusion models, allowing us to formulate sampling as stochastic optimization, where one can simply apply off-the-shelf solvers with lightweight iterates. Our experiments for various linear and nonlinear image restoration tasks demonstrate the strengths of our method compared with state-of-the-art sampling-based diffusion models. The code can be found at \href{https://github.com/NVlabs/RED-diff}{GitHub}.
\end{abstract}


\section{Introduction}
\vspace{-3mm}
Diffusion models such as Stable diffusion \citep{rombach2021highresolution} are becoming an integral part of nowadays visual foundation models. An important utility of such diffusion models is to use them as prior distribution for sampling in various downstream inverse problems appearing for instance in image restoration and rendering. This however demands samplers that are (i) {\it universal} and adaptive to various tasks without re-training for each individual task, and (ii) efficient and easy to tune.

There has been a few recent attempts to develop universal samplers for inverse problems; \citep{kawar2022denoising,song2023pseudoinverse,chung2022diffusion, kadkhodaie2021stochastic,graikos2022diffusion} to name a few. DDRM \citep{kawar2022denoising} was initially introduced to extend DDPM \citep{ho2020denoising} to handle linear inverse problems. It relies on SVD to integrate linear observations into the denoising process. DDRM however needs many measurements to work. Later on, $\Pi$GDM was introduced \citep{song2023pseudoinverse} to enhance DDRM. The crux of $\Pi$GDM is to augment the denoising diffusion score with the guidance from linear observations through inversion. In a similar vein, DPS \citep{chung2022diffusion} extends the score modification framework to general (nonlinear) inverse problems. The score modification methods in DPS and $\Pi$GDM, however, heavily resort to approximations. In essence, the nonlinear and recursive nature of the backward diffusion process renders the posterior distribution quite intractable and multimodal. However, DPS and $\Pi$GDM rely on a simple unimodal approximation of the score which is a quite loose approximation at many steps of the diffusion process. 

\begin{figure*}
    \centering
        \includegraphics[width=1.0\textwidth]{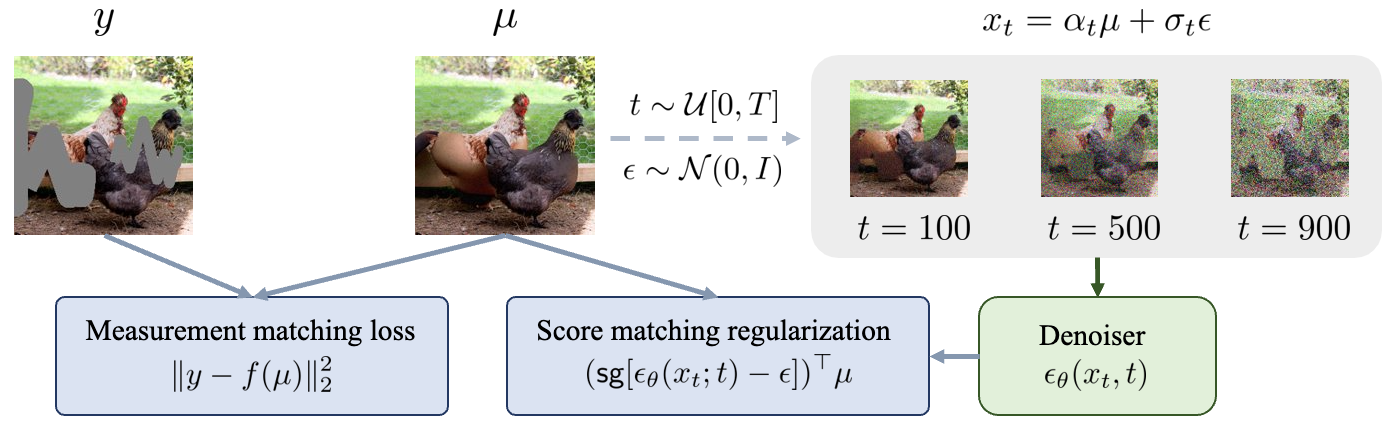}  
        \vspace{-4mm}
\caption{\small The schematic diagram of our proposed variational sampler (RED-diff). The forward denoising diffusion process gradually adds noise to the estimate $\mu$. The denoisers of the backward diffusion process apply score-matching regularization to the measurement matching loss. The refined estimate using optimization is then fed back to the forward process and the process repeats.}  \label{fig:diagram}
\end{figure*}

To sidestep the challenges for posterior score approximation, we put forth a fundamentally different approach based on variational inference \citep{blei2017variational, ahmed2012scalable, hoffman2013stochastic}. Adopting the denoising diffusion model as our data prior and representing the measurement model as a likelihood, we use variational inference to infer the posterior distribution of data given the observations. Our method essentially matches modes of data distribution with a Gaussian distribution using KL divergence. That leads to a simple (weighted) score-matching criterion that regularizes the measurement matching loss from observations via denoising diffusion process. Interestingly, the score-matching regularization admits an interpretable form with simple gradients; see Fig.~\ref{fig:diagram}. 

This resembles the regularization-by-denoising (RED) framework by \citet{romano2016the}, where denoisers at different stages of the diffusion process impose different structural constraints from high-level semantics to fine details. This is an important connection that views sampling as stochastic optimization. As a result, one can simply deploy the rich library of off-the-shelf optimizers for sampling which makes inference efficient, interpretable, and easy to tweak. We coin the term RED-diff to name our method. It is however worth noting that our framework differs from RED in several aspects: $(i)$ we derive our objective from a principled variational perspective that is well studied and understood, and $(ii)$ our regularization uses feedback from all the diffusion steps with different noise levels while RED uses a single denoising model.

For the success of the score matching regularization, denoisers at different timesteps need to be weighted properly. To do so, we propose a weighting mechanism based on densoing SNR at each timestep that upweights the earlier steps 
in the reverse diffusion process and down-weights the later timesteps. 
To verify the proposed idea, we conduct experiments and ablations for various linear and nonlinear inverse problems. Our main insights indicate that: $(i)$ RED-diff achieves superior image fidelity and perceptual quality compared with state-of-the-art samplers for image inverse problems; $(ii)$ RED-diff has lightweight iterates with no score Jacobian involved as in DPS and $\Pi$GDM, and as a result, it is more memory efficient and GPU friendly; $(iii)$ Our ablation studies suggest that the optimizer parameters such as learning rate and the number of steps are suitable knobs to tweak the trade-off between fidelity and perceptual quality.

\textbf{Contributions}. All in all, the main contributions of this paper are summarized as follows:
\begin{itemize}[noitemsep,nosep]

    \item We propose, RED-diff, a variational approach for general inverse problems, by introducing a rigorous maximum-likelihood framework that mitigates the posterior score approximation involved in recent $\Pi$GDM~\citep{song2023pseudoinverse} and DPS~\citep{chung2022diffusion}

    \item We establish a connection with regularization-by-denoising (RED) framework \citep{romano2016the}, which allows to treat sampling as stochastic optimization, and thus enables off-the-shelf optimizers for fast and tunable sampling
        
    \item We propose a weighting mechanism based on denoising SNR for the diffusion regularization
    
    \item We conduct extensive experiments for various linear and nonlinear inverse problems that show superior quality and GPU efficiency of RED-diff against state-of-the-art samplers such as $\Pi$GDM and DPS. Our ablations also suggest key insights about tweaking sampling and optimization to generate good samples.   
    
\end{itemize}

\section{Related works}
\vspace{-3mm}
Our work is primarily related to the following lines of work in the context of diffusion models.

\noindent\textbf{Diffusion models for inverse problems}:~There are several recent works to apply diffusion models in a plug-and-play fashion to inverse problems in various domains such as natural images \citep{kadkhodaie2021stochastic,jalal2021robust,kawar2022denoising,song2023pseudoinverse,chung2022diffusion,chung2022improving,graikos2022diffusion,chung2022parallel,chung2022solving}, medical images \citep{jalal2021robust}, and audio processing \citep{kong2020diffwave}. We primarily focus on images, where these works primarily differ in the way that they handle measurements. As some of the early works, \citet{kadkhodaie2021stochastic} and \citet{jalal2021robust} adopt Langevine dynamics for linear inverse problems and integrate the observation guidance via either projection \citep{kadkhodaie2021stochastic}, or gradient of the least-squares fidelity \citep{jalal2021robust}. Some other works adopt DDPM \citep{ho2020denoising} diffusion and alternate between diffusion denoising and projection steps \citep{choi2021ilvr,chung2021come}. The iterations however can accumulate error that pushes the trajectory off the prior manifold, and thus MCG method \citep{chung2022improving} proposes an additional correction term inspired by the manifold constraint to keep the iterations close to the manifold. DDRM \citep{kawar2022denoising} extends DDPM to solve linear inverse problems using matrix SVD, but it fails when the number of measurements is small.  

To address this shortcoming, recent methods aim to provide guidance by differentiating through the diffusion model in the form of reconstruction guidance~\citep{ho2022video}, which is further extended in DPS \citep{chung2022diffusion} to nonlinear inverse problems. $\Pi$GDM~\citep{song2023pseudoinverse} introduces pseudoinverse guidance that improves the guidance approximation by inverting the measurement model. Its scope is however limited to linear and certain semi-linear tasks (such as JPEG~\citep{kawar2022jpeg}). However, both $\Pi$GDM and DPS heavily rely on an approximation of the intractable posterior score, which is quite crude for non-small noise levels at many steps of the diffusion process. Note also that, a different method has also been recently proposed by \citet{graikos2022diffusion}, which regularizes the reconstruction term of inverse problems with the diffusion error loss. This is similar to the traditional plug-and-play prior (P$^3$) approach for inverse problems \citep{venkatakrishnan2013plug} that roots back to ADMM optimization \citep{boyd2011distributed}. Our method is however closer in spirit to the RED framework, which is known to be fundamentally different from P$^3$ framework, and offers more flexibility for optimizer and tuning; see e.g., \citep{romano2016the, cohen2021regularization}.










\noindent\textbf{Diffusion models for 3D}: A few recent works have adopted distillation loss optimization to generate 3D data from 2D diffusion priors, which is related to our view of treating sampling as optimization. For instance, DreamFusion \citep{poole2022dreamfusion} and ProfilicDreamer \citep{wang2023prolificdreamer} adopt a probability density distillation loss as the criterion for text-to-3D generation. Followup works include SparseFusion \citep{zhou2022sparsefusion} that generates 3D given a few (e.g. just two) segmented
input images with known relative pose, and NeuralLift-360 \citep{xu2022neurallift} that lifts a single 2D image to 3D. All these methods use a distillation loss, that bears resemblance with our (unweighted) denoising regularization. However, they aim to optimize for a parametric 3D NeRF model that is fundamentally different from our goal.

\section{Background}
\label{sec:problem_statement}
\vspace{-3mm}
In this section, we first review diffusion models in Section~\ref{sec:diffusion_process} and we discuss how they are used for solving inverse problems in Section~\ref{sec:var_infer}.

\vspace{-2mm}
\subsection{Denoising diffusion models}
\label{sec:diffusion_process}
\vspace{-3mm}
Diffusion models~\citep{sohl2015deep,ho2020denoising,song2020score} consist of two processes: a forward process that gradually adds noise to input images and a reverse process that learns to generate images by iterative denoising. Formally the forward process can be expressed by the variance preserving stochastic differential equation (VP-SDE)~\citep{song2020score} $dx = -\frac{1}{2} \beta(t) x dt + \sqrt{\beta(t)} dw$ for $t \in [0, T]$ where $\beta(t):=\beta_\text{min} + (\beta_\text{max} - \beta_\text{min})\frac{t}{T}$ rescales the time variable, and $dw$ is the standard Wiener process. The forward process is designed such that the distribution of $x_T$ at the end of the process converges to a standard Gaussian distribution (i.e., $x_T \sim \mathcal{N}(0, I)$). The reverse process is defined by $dx = -\frac{1}{2}\beta(t) x dt - \beta(t) \nabla_{x_t} \log p(x_t)+ \sqrt{\beta(t)} d\bar{w}$ where $\nabla_{x_t} \log p(x_t)$ is \textit{the score function} of diffused data at time $t$, and $d\bar{w}$ is the reverse standard Wiener process. 

Solving the reverse generative process requires estimating the score function. In practice, this is done by sampling from the forward diffusion process and training the score function using the denoising score-matching objective~\citep{vincent2011connection}. Specifically, diffused samples are generated by:%
\begin{align}
    x_t = \alpha_t x_0 + \sigma_t \epsilon,  \quad \epsilon \sim \mathcal{N}(0,I), \quad t \in [0, T] \label{eq:forward-process}
\end{align}
where $x_0 \sim p_{\text{data}}$ is drawn from data distribution, $\sigma_t = 1-e^{-\int_0^t \beta(s) ds}$, and $\alpha_t=\sqrt{1-\sigma_t^2}$. Let's denote the parameterized score function (i.e., diffusion model) by $\epsilon_{\theta}(x_t;t) \approx -\sigma_t \nabla_{x_t} \log p(x_t)$ with parameters $\theta$, we can train $\epsilon_{\theta}(x_t;t)$ with a mixture of Euclidean losses, such as
\begin{align*}
    \min_{\theta} \mathbb{E}_{x_0 \sim p_{\text{data}}(x_0), \epsilon \sim \mathcal{N}(0,I), t \sim \mathcal{U}[0,T] }\left[||\epsilon -  \epsilon_{\theta}(x_t;t) ||_2^2 \right].
\end{align*}
Other loss-weighting functions for $t$ can be used as well.
Given a trained score function, samples can be generated using DDPM \citep{ho2020denoising}, DDIM \citep{song2020denoising}, or other solvers~\citep{lu2022dpm, zhang2022deis,dockhorn2022genie}.

\subsection{Score approximation for inverse problems}
\label{sec:var_infer}
\vspace{-2mm}
An inverse problem is often formulated as finding $x_0$ from a (nonlinear and noisy) observation:
\begin{align} \label{eq:meassurement}
    y=f(x_0) + v,  \quad v \sim \mathcal{N}(0, \sigma_v^2I)
\end{align}
where the forward (a.k.a measurement) model $f$ is known. In many applications, such as inpainting, this is a severely ill-posed task that requires a strong prior to find a plausible solution. Our goal is to leverage the prior offered by (pretrained) diffusion models, in a plug-and-play fashion, to efficiently sample from the conditional posterior. Let's denote the prior distributions imposed by diffusion models as $p(x_0)$. The measurement models can be represented by $p(y|x_0) := \mathcal{N}(f(x_0), \sigma_v^2)$. The goal of solving inverse problems is to sample from the posterior distribution $p(x_0|y)$.

As we discussed in the previous section, diffusion models rely on the estimated score function to generate samples. In the presence of the measurements $y$, they can be used for generating plausible $x_0 \sim p(x_0|y)$ as long as an approximation of the conditional score for $p(x_t|y)$ over all diffusion steps is available.
This is the idea behind $\Pi$GDM~\citep{song2023pseudoinverse} and DPS~\citep{chung2022diffusion}. 
Specifically, the conditional score for $p(x_t|y)$ based on Bayes rule is simply obtained as 
\begin{align}
    \nabla_x \log p(x_t|y) = \nabla_x \log p(y|x_t) + \nabla_x \log p(x_t)
\end{align}
The overall score is a superposition of the model likelihood and the prior score. 
While $\nabla_x \log p(x_t)$ is easily obtained from a pretrained diffusion model, the likelihood score is quite challenging and intractable to estimate without any task-specific training.
This can be seen from the fact that $p(y|x_t) = \int   p(y|x_0) p(x_0|x_t) dx_0$. Although $p(y|x_0)$ takes a simple Gaussian form, the denoising distribution $p(x_0|x_t)$ can be highly complex and multimodal~\citep{xiao2022DDGAN}. As a result, $p(y|x_t)$ can be also highly complex. To sidestep this, prior works \citep{song2023pseudoinverse,chung2022diffusion,kadkhodaie2021stochastic,ho2022video} resort to a unimodal approximation of $p(x_0|x_t)$ using either a simple Gaussian assumption, or the MMSE estimate that states:
\begin{align}
    \mathbb{E}[x_0 | x_t] = \frac{1}{\alpha_t} (x_t - \sigma_t \epsilon_{\theta}(x_t,t)).
\end{align}

\section{Variational diffusion sampling}
\vspace{-3mm}
In this section, we introduce our variational perspective on solving inverse problems.
To cope with the shortcomings of previous methods for sampling the conditional posterior $p(x_0|y)$, we propose a variational approach based on KL minimization
\begin{align}
    \min_q KL\big(q(x_0|y) || p(x_0|y)\big)    \label{eq:kl_loss}
\end{align}
where $q := \mathcal{N}(\mu, \sigma^2 I)$ is a variational distribution. The distribution $q$ seeks the dominant mode in the data distribution that matches the observations.  
It is easy to show that the KL objective in Eq.~\ref{eq:kl_loss} can be expanded as
\begin{align} 
    &KL\big(q(x_0|y) \| p(x_0|y)\big) = \underbrace{-\mathbbm{E}_{q(x_0|y)}\big[\log p(y|x_0)\big]\!+\!KL\big(q(x_0|y) \| p(x_0)\big)}_{\text{term (i)}} + \underbrace{\log p(y)}_{\text{term (ii)}} \label{eq:var_bound}
\end{align}
where term (i) is the variational bound that is often used for training variational autoencoders~\citep{kingma2013auto,rezende2014stochastic} and term (ii) is the observation likelihood that is \textit{constant} w.r.t.~$q$. Thus, to minimize the KL divergence shown in Eq.~\ref{eq:kl_loss} w.r.t.~$q$, it suffices to minimize the variational bound (term (i)) in Eq.~\ref{eq:var_bound} w.r.t.~$q$. This brings us to the next claim.

\textbf{Proposition 1}.~{\it The KL minimization w.r.t~$q$ in Eq.~\ref{eq:kl_loss} is equivalent to minimizing the variational bound (term (i) in Eq.~\ref{eq:var_bound}), that itself obeys the score matching loss: }
\begin{align}
    &\min_{\{\mu, \sigma\}} \ \mathbb{E}_{q(x_0|y)}\left[\frac{\|y - f(x_0)\|_2^2}{2\sigma_v^2}\right] + \int_{0}^{T}\! \tilde{\omega}(t) \mathbbm{E}_{q(x_t|y)} \Big[\big\|\nabla_{x_t}\!\log q(x_t|y) - \nabla_{x_t}\!\log p(x_t)\big\|^2_2 \Big] dt, \label{eq:prop1}
\end{align}
{\it where $q(x_t | y) = \mathcal{N}(\alpha_t \mu, (\alpha_t^2 \sigma^2 + \sigma_t^2)I)$ produces samples $x_t$ by drawing $x_0$ from $q(x_0 | y)$ and applying the forward process in Eq.~\ref{eq:forward-process}, and $\tilde{\omega}(t)=\beta(t)/2$ is a loss-weighting term.}

Above, the first term is the measurement matching loss (i.e., reconstruction loss) obtained by the definition of $p(y|x_0)$, while the second term is obtained by expanding the KL term in terms of the score-matching objective as shown in \citep{vahdat2021score, song2021maximum}, and $\tilde{\omega}(t) = \beta(t)/2$ is a weighting based on maximum likelihood (the proof is provided in the supplementary material). The second term can be considered as a score-matching regularization term imposed by the diffusion prior. The integral is evaluated on a diffused trajectory, namely $x_t \sim q(x_t | y)$ for $t \in [0, T]$, which is the forward diffusion process applied to $q(x_0 | y)$. Since $q(x_0 | y)$ admits a simple Gaussian form, we can show that $q(x_t|y)$ is also a Gaussian in the form $q(x_t|y) = \mathcal{N}(\alpha_t \mu, (\alpha_t^2 \sigma^2 + \sigma_t^2)I)$ (see \citep{vahdat2021score}). Thus, the score function $\nabla_{x_t}\!\log q(x_t|y)$ can be computed analytically.

Assuming that the variance of the variational distribution is a small constant value near zero (i.e., $\sigma \approx 0$), the optimization problem in Eq.~\ref{eq:prop1} can be further simplified to:
\begin{align}\label{eq:prop1_sim}
    \min_{\mu}\ \underbrace{\|y - f(\mu)\|^2}_{\text{recon}} + \underbrace{\mathbbm{E}_{t,\epsilon} \big[2\omega(t) (\sigma_v/\sigma_t)^2 ||\epsilon_{\theta}(x_t;t) - \epsilon||_2^2 \big]}_{\text{reg}},
\end{align}
where $x_t = \alpha_t \mu + \sigma_t \epsilon$.
In a nutshell, solving the optimization problem above will find an image $\mu$ that reconstructs the observation $y$ given the measurement model $f$, while having a high likelihood under the prior as imposed by the regularization term. 


\noindent\textbf{Remark [Noiseless observations].}~If the observation noise $\sigma_v=0$, then from \eqref{eq:var_bound} the reconstruction term boils down to a hard constraint which can be represented as an indicator function $\mathbbm{1}_{\{y=f(\mu)\}}$ that is zero when $y=f(\mu)$ and infinity elsewhere. In practice, however we can still use \eqref{eq:prop1} with a small $\sigma_v$ as an approximation.

\subsection{Sampling as stochastic optimization}
\label{sec:sampling_optimization}
\vspace{-2mm}
The regularized score matching objective Eq.~\ref{eq:prop1_sim} allows us to formulate sampling as optimization for inverse problems. In essence, the ensemble loss over different diffusion steps advocates for stochastic optimization as a suitable sampling strategy. 

However, in practice the choice of weighting term $\tilde{\omega}(t)$ plays a key role in the success of this optimization problem. Several prior works on training diffusion models~\citep{ho2020denoising,vahdat2021score,Karras2022edm,choi2022perception} have found that reweighting the objective over $t$ plays a key role in trading content vs. detail at different diffusion steps which we also observe in our case (more information in Section~\ref{subsec:tune_weights}). Additionally, the second term in Eq.~\ref{eq:prop1_sim} marked by ``reg'' requires backpropagating through pretrained score function which can make the optimization slow and unstable. Next, we consider a generic weighting mechanism $\tilde{\omega}(t)=\beta(t)\omega(t)/2$ for a positive-valued function $\omega(t)$, and we show that if the weighting is selected such that $\omega(0)=0$, the gradient of the regularization term can be computed efficiently without backpropagating through the pretrained score function.


\textbf{Proposition 2}.~{\it If $\omega(0)=0$ and $\sigma=0$, then the gradient of the score matching regularization loss admits}
\begin{align*}
    \nabla_{\mu} {\rm reg}(\mu) & =  \mathbbm{E}_{t \sim \mathcal{U}[0,T],\epsilon \sim \mathcal{N}(0,I)} \big[ \lambda_t (\epsilon_{\theta}(x_t;t) - \epsilon) \big]
\end{align*}
{\it where $\lambda_t:=\frac{2T\sigma_v^2 \alpha_t}{\sigma_t}\frac{d \omega(t)}{dt}$.} 



%
%
\noindent\textbf{First-order stochastic optimizers}.~Based on the simple expression for the gradient of score-matching regularization in Proposition 2, we can treat time as a uniform random variable. Thus by sampling randomly over time and noise, we can easily obtain unbiased estimates of the gradients. Accordingly, first-order stochastic optimization methods can be applied to search for $\mu$. We list the iterates under Algorithm 1. Note that we define the loss per timestep based on the instantaneous gradient, which can be treated as a gradient of a linear loss. We introduce the notation ($\mathsf{sg}$) as stropped-gradient to emphasize that score  is not differentiated during the optimization. The ablations in Section~\ref{subsec: ablate_sampling} show that (descending) time stepping from $t=T$ to $t=0$, as in standard backward diffusion samplers such as DDPM and DDIM, performs better than random time sampling in practice.

Note that Proposition 2 derives the gradient for the case with no dispersion (i.e., $\sigma=0$) for simplicity. The extension to nonzero dispersion is deferred to the appendix.

\subsection{Regularization by denoising}
\label{subsec:red}
\vspace{-2mm}
Note that our variational sampler strikes resemblance with the regularization by denoising (RED) framework \citep{romano2016the}. In essence, RED is a flexible way to harness a given denoising engine for treating general inverse problems. RED regularization effectively promotes smoothness for the image according to some image-adaptive Laplacian prior. To better understand the connection with RED, let us look at the loss per timestep of our proposed variational sampler. From the gradient expression in Proposition 2, we can form the loss at timestep $t$ as
\begin{align} 
    \|y - f(\mu)\|^2 + \lambda_t ( {\mathsf{sg}}[\epsilon_{\theta}(x_t;t) - \epsilon] )^{\top} \mu  \label{eq:inst_loss}
\end{align}
Interestingly, the regularization term is similar to RED. A small regularization term implies that either the diffusion reaches the fixed point, namely $\epsilon_{\theta}(x_t;t) = \epsilon$, or the residual only contains noise with no contribution left from the image. Note also that again similar to RED, the gradient of the regularizer is quite simple and tractable. Of course, there are fundamental differences with RED including the generative nature of our diffusion prior, and the main fact that we use the entire diffusion trajectory for regularization. However, given these differences, we believe this is an important connection to leverage RED utilities for improved sampling of diffusion models in inverse problems. It is also worth commenting that the earlier work by \citet{reehorst2018regularization} also draws connections between RED and score matching based on a single denoiser.

\begin{algorithm}[t!]
\small
\caption{Variational sampler (RED-diff)}\label{alg:varsgd-alg}
\begin{algorithmic}
\INPUT $y$, $f(\cdot)$, $\sigma_v$, $L$, $\{\alpha_t, \sigma_t, \lambda_t\}_{t=1}^T$
\Initialize $\mu_0$
\For {$\ell = 1,\ldots,L$}
    \State $t \sim \mathcal{U}[0,T]$
    \State $\epsilon \sim \mathcal{N}(0, I_n)$
    \State $x_t=\alpha_t \mu + \sigma_t \epsilon$
    \State $loss=\|y - f(\mu)\|^2 + \lambda_t ({\mathsf{sg}}[\epsilon_{\theta}(x_t;t) - \epsilon] )^{\top} \mu$
    \State  $\mu \leftarrow \mathsf{OptimizerStep}(loss)$
\EndFor
\RETURN $\mu$
\end{algorithmic}
\end{algorithm}



\subsection{Weighting mechanism}
\label{subsec:tune_weights}
\vspace{-2mm}

In principle, timestep weighting plays a key role in training diffusion models. Different timesteps are responsible for generating different structures ranging from large-scale content in the last timesteps to fine-scale details in the earlier timesteps~\citep{choi2022perception}. For effective regularization, it is thus critical to properly tune the denoiser weights $\{\lambda_t\}$ in our Algorithm~\ref{alg:varsgd-alg}. We observed that the regularization term in Eq.~\ref{eq:inst_loss} is sensitive to noise schedule. For example, in the variance-preserving scenario, it drastically bellows up as $t$ approaches zero; \textcolor{red}.


To mitigate the regularization sensitivity to weights, it is more desirable to define the regularization in the signal domain, that is compatible with the fitting term as
\begin{align}
    \|y - f(\mu)\|^2 + \lambda (\mathsf{sg}[\mu - \hat{\mu}_t])^{\top}\mu,  \label{eq:inst_loss_img}
\end{align}
where $\lambda$ is a hyperparameter that balances between the prior and likelihood and $\hat{\mu}_t$ is the MMSE predictor of clean data.
Here, we want the constant $\lambda$ to control the trade-off between bias (fit to observations) and variance (fit to prior). In order to come up with the interpretable loss in \eqref{eq:inst_loss_img}, one needs to rescale the noise residual term $\epsilon_{\theta}(x_t;t) - \epsilon$.

Recall that the denoiser at time $t$ observes $x_t=\alpha_t x_0 + \sigma_t \epsilon$. MMSE estimator also provides denoising as
\begin{align}
    \hat{\mu}_t = \mathbb{E}[\mu | x_t] = \frac{1}{\alpha_t} (x_t - \sigma_t \epsilon_{\theta}(x_t;t)). 
\end{align}
Thus, one can show that
\begin{align*}
   \mu - \hat{\mu}_t  = (\sigma_t/\alpha_t) (\epsilon_{\theta}(x_t;t) - \epsilon)
\end{align*}
where we define ${\rm SNR}_t:=\alpha_t/\sigma_t$ as the signal-to-noise ratio. Accordingly, by choosing $\lambda_t=\lambda/{\rm SNR}_t$, we can simply convert the noise prediction formulation in \eqref{eq:inst_loss} to clean data formulation in \eqref{eq:inst_loss_img}.




\section{Experiments}
\label{sec:exps}
\vspace{-2mm}
In this section, we compare our proposed variational approach, so termed RED-diff, against recent state-of-the-art techniques for solving inverse problems on different image restoration tasks. For the prior, we adopt publicly available checkpoints from the guided diffusion model\footnote{\url{https://github.com/openai/guided-diffusion}} that is pretrained based on $256 \times 256$ ImageNet~\citep{russakovsky2015imagenet}; see details in the appendix. We consider the unconditional version. For the proof of concept, we report findings for various linear and nonlinear image restoration tasks for a 1k subset of ImageNet~\citep{russakovsky2015imagenet} validation dataset\footnote{\url{https://bit.ly/eval-pix2pix}}. Due to space limitation, we defer more elaborate experiments and ablations to the appendix. Next, we aim to address the following important questions:

\begin{itemize}
    \item How does the proposed variational sampler (RED-diff) compare with state-of-the-art methods such as DPS, $\Pi$GDM, and DDRM in terms of quality and speed? 
    \vspace{-1mm}
    \item What is a proper sampling strategy and weight-tuning mechanism for the variational sampler?
\end{itemize}

\vspace{-2mm}
\noindent\textbf{Sampling setup}. We adopt linear schedule for $\beta_t$ from $0.0001$ to $0.02$ for $1,000$ timesteps. For simplicity, we always use uniform spacing when we iterate the timestep. For our variational sampler we adopt Adam optimizer with $1,000$ steps, and set the momentum pair $(0.9,0.99)$ and initial learning rate $0.1$. No weight decay regularization is used. The optimizer is initialized with the degraded image input. We also choose descending time stepping from $t=T$ to $t=1$ as demonstrated by the ablations later in Section~\ref{sec:timestep_sampling}. Across all methods, we also use a batch size of 10 using RTX 6000 Ada GPU with 48GB RAM.

\begin{figure}
    \centering
        \hspace{0mm}\includegraphics[width=0.85\textwidth]{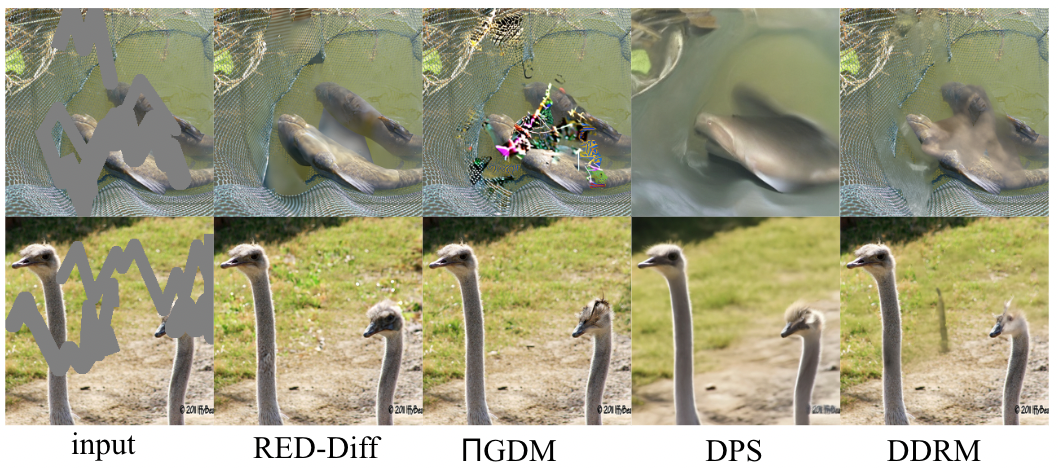} 
        \vspace{-5mm}
\caption{\small Comparison of the proposed variational sampler with alternatives for inpainting representative ImageNet examples. Each sampler is tuned for the best performance. }  \label{fig:inp_comp}
\end{figure}

\noindent\textbf{Comparison}. For comparison we choose state-of-the-art techniques including DPS \citep{chung2022diffusion}, $\Pi$GDM \citep{song2023pseudoinverse}, and DDRM \citep{kawar2022denoising} as existing alternatives for sampling inverse problems. We tune the hyper-parameters as follows:
\begin{itemize}[noitemsep,nosep]
    \item DPS \citep{chung2022diffusion}: $1,000$ diffusion steps, tuned $\eta=0.5$ for the best performance;
    \item $\Pi$GDM \citep{song2023pseudoinverse}: $100$ diffusion steps, tuned $\eta=1.0$, and observed that $\Pi$GDM does perform worse for $1000$ steps;
    \item DDRM \citep{kawar2022denoising}: tested for both $20$ and $100$ steps, and set $\eta=0.85$, $\eta_b=1.0$. DDRM is originally optimized for 20 steps.
\end{itemize}
For evaluation, we report metrics including Kernel Inception Distance (KID,~\citep{binkowski2018demystifying}), LPIPS, SSIM, PSNR, and top-1 Classifier Accuracy (CA) of a pre-trained ResNet50 model~\citep{he2015deep}. 

\vspace{-2mm}
\subsection{Image inpainting}
\vspace{-2mm}
For inpainting evaluation, we adopt $1$k samples from the ImageNet dataset and random masks from Palette \citep{saharia2022palette}. We tune $\lambda=0.25$ for the SNR-based denoiser weight tuning discussed in Section~\ref{subsec:tune_weights}. A few representative examples are shown in Fig.~\ref{fig:inp_comp}. For a fair comparison, we choose a relatively hard example in the first row, and an easier one in the bottom row. It is evident that RED-diff identifies the context, and adds the missing content with fine details. $\Pi$GDM however fails to inpaint the hard example, and DPS and DDRM inpaint blurry contents. More examples are provided in the appendix.

Quantitative results are also listed in Table \ref{tab:eval_inp}. One can see that RED-diff consistently outperforms the alternative samplers across all metrics such as KID and PSNR with a significant margin. This indicates not only more faithful restoration by RED-diff but also better perceptual quality images compared with alternative samplers.

Finally, note that RED-diff iterations are quite lightweight with only forward passing to the diffusion score network. In contrast, DPS and $\Pi$GDM require score network inversion by differentiating through the diffusion denoisers. This in turn is a source of instability and renders the steps computationally expensive. Likewise, DDRM involves SVD calculations that are costly. We empirically validate these by comparing the time per-step and GPU memory usage in Table \ref{tab:eval_inp}.

\begin{table}
    \centering
    \renewcommand\tabcolsep{5pt}
    \caption{\small Performance of different samplers for ImageNet inpainting using pretrained unconditional guided diffusion model. For RED-diff we set $lr=0.5$. For time per step (for each sample) we use the maximum batch size that fits the GPU memory. All methods run on a single NVIDIA RTX 6000 Ada GPU with 48GB RAM.}
    \vspace{2mm}
    \scalebox{0.92}{
    \centering
    \begin{tabular}{l|ccccc|c|c}
    Sampler & PSNR(dB) $\uparrow$ & SSIM $\uparrow$ & KID $\downarrow$ & LPIPS $\downarrow$  & top-1 $\uparrow$ & time per step (sec) $\downarrow$ & max batch size $\uparrow$  \\
    \hline
    \hline
    \rule{0pt}{2ex} 
    DPS & 21.27 & 0.67 &  15.28 & 0.26 & 58.2 & 0.23 & 15   \\
    $\Pi$GDM & 20.30 & 0.82 &  4.50 & 0.12 & 67.8  & 0.24 & 15  \\
    DDRM & 20.72 & 0.83 &  2.5 & 0.14 & 68.6 & 0.1  & 25   \\
    RED-diff & \bf{23.29} & \bf{0.87} & \bf{0.86} & \bf{0.1} & \bf{72.0}  & \bf{0.05} & \bf{30}  \\
    \hline
    \end{tabular}
    }
    \label{tab:eval_inp}
    \vspace{-0.2cm}
\end{table}

    
    

    
    

\vspace{-2mm}
\subsection{Nonlinear inverse problems}
\vspace{-2mm}
For various nonlinear inverse problems we assess RED-diff on ImageNet data. We choose DPS as the baseline since $\Pi$GDM and DDRM only deal with linear inverse problems.


\begin{table*}
    \centering
    \renewcommand\tabcolsep{5pt} 
    \caption{Performance of different samplers for nonlinear tasks based on ImageNet data.}
    \begin{tabular}{l||cc|cc|cc}
    \toprule
    \textbf{Task} & \multicolumn{2}{c}{\textbf{Phase Retrieval}} & \multicolumn{2}{c}{\textbf{HDR}} & \multicolumn{2}{c}{\textbf{Deblurring}} \\
    \cmidrule(r){1-1} \cmidrule(lr){2-3} \cmidrule(lr){4-5} \cmidrule(l){6-7}
    Metrics & DPS & RED-diff & DPS & RED-diff & DPS & RED-diff \\
    \midrule
    PSNR(dB) $\uparrow$ & 9.99 & \textbf{10.53} & 7.94 & \textbf{25.23} & 6.4 & \textbf{45.00} \\
    SSIM $\uparrow$ & 0.12 & \textbf{0.17} & 0.21 & \textbf{0.79} & 0.19 & \textbf{0.987} \\
    KID $\downarrow$ & \textbf{93.2} & 114.0 & 272.5 & \textbf{1.2} & 342.0 & \textbf{0.1} \\
    LPIPS $\downarrow$ & 0.66 & \textbf{0.6} & 0.72 & \textbf{0.1} & 0.73 & \textbf{0.0012} \\
    top-1 $\uparrow$ & 1.5 & \textbf{7.2} & 4.0 & \textbf{68.5} & 6.4 & \textbf{75.4} \\
    \bottomrule
    \end{tabular}
    \label{tab:nonlinear_results}
\end{table*}

\noindent\textbf{High dynamic range (HDR)}. We choose the nolinear HDR task as a candidate to verify RED-diff. HDR performs the clipping function $f(x) = {\rm clip}(2x, -1, 1)$ on the normalized RGB pixels. Again, we choose $\lambda=0.25$ and $lr=0.5$, and 100 steps. For DPS we choose $\zeta_i = \frac{0.1}{\|y - A(\hat{x}_0(x_i))\|}$ after grid search over the nominator. While RED-diff converges to good solutions, DPS struggles to find a decent solution even after tuning. The metrics listed under Table \ref{tab:nonlinear_results} demonstrate the gap.

\noindent\textbf{Phase retrieval}. We test on phase retrieval task as well. The task deals with reconstructing the phase from only magnitude observations in the Fourier domain. It is a difficult task especially for ImageNet dataset with diverse details and structures. Again, for RED-diff we use the weight $\lambda=0.25$ and $lr=0.5$, while for DPS we optimize the step size $\zeta_i = \frac{0.4}{\|y - A(\hat{x}_0(x_i))\|}$. While both methods face with challenges for recovering faithful images, RED-diff performs better and achieves higher scores for most of the metrics; see Table \ref{tab:nonlinear_results}.

\noindent\textbf{Deblurring}. We also test another nonlinear scenario that deals with nonlinear deblurring. We adopt the same setup as in DPS \citep{chung2022diffusion} with the blur kernel adopted from a pretrained UNet. For RED-diff we choose $\lambda=0.25$ and $lr=0.5$. For DPS also after grid search over the coefficients we end up with $\zeta_i = \frac{1.0}{\|y - A(\hat{x}_0(x_i))\|}$. DPS struggles for this nonlinear tasks. In general, DPS is sensitive to step size and initialization, while RED-diff is not sensitive and achieves much better scores as listed in Table \ref{tab:nonlinear_results}.

\vspace{-2mm}
\subsection{Ablations}
\vspace{-2mm}
We provide ablations to verify the role of different design components in the proposed algorithm such as denoiser weight tuning, and sampling strategy.

\subsubsection{Denoiser weighting mechanism}
\begin{figure}
    \centering
        \includegraphics[width=0.75\textwidth]{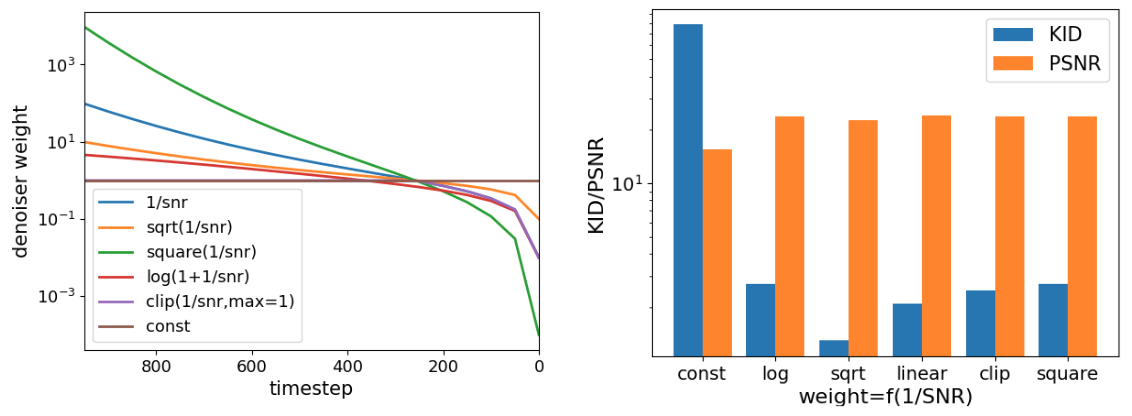}  
        \vspace{-5mm}
\caption{\small Ablation for denoiser weight tuning. Left: denoiser weight over timesteps (reversed); right: KID and PSNR vs. $\lambda$ for different monotonic functions of inverse SNR.}  \label{fig:ablate_weight}
\end{figure}

\vspace{-2mm}
As discussed in Section~\ref{subsec:red}, the variational sampler resembles regularization by the denoising diffusion process. When sampling in descending order, namely from $t=T$ to $t=1$, each denoiser regularizes different structures from high-level semantics to low-level fine details, respectively. To effect prior at different image scales, each denoiser needs to be tuned properly. We proposed inverse SNR (i.e., $1/{\rm SNR}_t$) as the base weight per timestep in Section~\ref{subsec:tune_weights}. To validate that choice, we ablate different monotonic functions of SNR to weight denoisers over time. The weights are plotted in Fig.~\ref{fig:ablate_weight} (left) over timesteps. The corresponding KID and PSNR metrics are also shown in Fig.~\ref{fig:ablate_weight} (right) for Platte inpainting for different weighting mechanisms. It is observed that the square root decay (i.e., $1/\sqrt{{\rm SNR}}$) and linear mechanism (i.e., $1/{\rm SNR}$) are the best strategies for KID and PSNR, respectively.




\subsubsection{Timestep sampling}
\label{sec:timestep_sampling}
\vspace{-2mm}
We consider five different strategies when sampling the timestep $t$ during optimization, namely: (1) random sampling; (2) ascending; (3) descending; (4) min-batch random sampling; and (5) mini-batch descending sampling. We adopt Adam optimizer with $1,000$ steps and choose the linear weighting mechanism with $\lambda=0.25$. Random sampling (1) uniformly selects a timestep $t \in [1,T]$, while ascending and descending sampling are ordered over timesteps. It is seen that descending sampling performs significantly better than others. It starts from the denoiser at time $t=T$, adding semantic structures initially, and then fine details are gradually added in the process. This appears to generate images with high fidelity and perceptual quality. We also tested batch sampling with $25$ denoisers per iteration, for $40$ iterations. It is observed that batch sampling smears the fine texture details. See appendix for more details.

\section{Conclusions and limitations}
\vspace{-2mm}
This paper focuses on the universal sampling of inverse problems based on diffusion priors. It introduces a variational sampler, termed RED-diff, that naturally promotes regularization by the denoising diffusion process (DDP). Denoisers at different steps of DDP concurrently impose structural constraints from high-level semantics to low-level fine details. To properly tune the regularization, we propose a weighting mechanism based on denoising SNR. Our novel perspective views sampling as stochastic optimization that embraces off-the-shelf optimizers for efficient and tunable sampling. Our experiments for several image restoration tasks exhibit the strong performance of RED-diff compared with state-of-the-art alternatives for inverse problems. 

One of the limitations of our variational sampler pertains to the lack of diversity. It is mode-seeking in nature and promotes MAP solutions. We will investigate methods that encourage diversity e.g., by tuning the optimizer, introducing more expressively in the variational distribution, or modifying the criterion by adding dispersion terms as in Stein variational gradient descent~\cite{liu2016stein}. Additionally, more extensive experiments for 3D generation tasks will solidify the merits of our variational sampler.

\newpage
\bibliography{refs}
\bibliographystyle{iclr2024_conference}

\appendix
\onecolumn

\section{Proofs and Derivations}
\subsection{Proof of Proposition 1}
As discussed in section 4, using Bayes rule, one can re-write the KL objective in \eqref{eq:kl_loss} as
\begin{align*}
    &KL\big(q(x_0|y) \| p(x_0|y)\big)  = \underbrace{-\mathbbm{E}_{q(x_0|y)}\big[\log p(y|x_0)\big] + KL\big(q(x_0|y) \| p(x_0)\big)}_{\rm term(i)} + \underbrace{\log p(y)}_{{\rm term(ii)}}
\end{align*}
For minimization purposes, term(ii) is a constant, and we can ignore it. The term(i) however has also two parts. The first part is simply the reconstruction loss. Based on the measurement model in \eqref{eq:meassurement} of the main paper, since we assumed noise is i.i.d. Gaussian, the first part is simply derived as
\begin{align*}
    \mathbbm{E}_{q(x_0|y)}[\log p(y|x_0)] = - \frac{1}{2\sigma_v^2} \mathbbm{E}_{q(x_0|y)}[\|y - f(x_0)\|^2]
\end{align*}

Using theorem 1 in \cite{song2021maximum}, assuming some mild assumptions on the growth of $\log q(x_t|y)$ and $p(x_t)$ at infinity, we have
\begin{align}
    &KL(q(x_0|y) \| p(x_0)))   = \int_{0}^{T} \frac{\beta(t)}{2} \mathbbm{E}_{q(x_t|y)} \Big[\big\|\nabla_{x_t} \log q(x_t|y) - \nabla_{x_t} \log p(x_t)\big\|^2 \Big] dt   \label{eq:score_match_loss}
\end{align}
over the denoising diffusion trajectory $\{x_t\}$ for positive values $\{\beta(t)\}$. This essentially implies that a weighted score-matching over the continuous denoising diffusion trajectory is equal to the KL divergence.
In practice, however, we are often interested in a reweighting of r.h.s. in Eq.~\ref{eq:score_match_loss} that leads to other divergence measures \cite{song2021maximum}.

\subsection{Proof of Proposition 2}
The regularization term is essentially the score matching loss in the r.h.s. of \eqref{eq:score_match_loss}. In practice, we often use a weighting scheme different than $\beta(t)/2$ that corresponds to maximum likelihood estimation \cite{song2021maximum}. For re-weighting, it is useful to recognize the following Lemma, which leverages integration by part, with the complete proof provided in \cite{song2021maximum}.

\noindent\textbf{Lemma 1}. {\it The time-derivative of the KL divergence at timestep $t$ obeys}
\begin{align}
    \frac{d KL\big(q(x_t|y) \| p(x_t)\big)}{d t} = -\frac{\beta(t)}{2} \mathbbm{E}_{q(x_t|y)} \Big[\big\|\nabla_{x_t} \log q(x_t|y) - \nabla_{x_t} \log p(x_t)\big\|^2 \Big]  \label{eq:kl_time}
\end{align}

This Lemma is intuitive based on \eqref{eq:score_match_loss}. Now, to allow a generic weighting, one can simply weight \eqref{eq:kl_time} with $\omega(t)$, Then, under the condition $\omega(0)=0$, the r.h.s. of \eqref{eq:score_match_loss} can be written as (see e.g., \cite{song2021maximum})
\begin{align*}
       \int_{0}^{T} \frac{\beta(t)}{2} \omega(t) \mathbbm{E}_{q(x_t|y)} \Big[\big\|\nabla_{x_t} & \log  q(x_t|y) - \nabla_{x_t} \log p(x_t)\big\|^2 \Big] dt \\ 
       &=  - \int_{0}^{T} \omega(t) \frac{d KL\big(q(x_t|y) \| p(x_t)\big)}{d t} dt \\
    & \stackrel{(a)}{=} \underbrace{- \omega(t) KL\big(q(x_t|y) \| p(x_t)\big) \Big]_{0}^{T} }_{=0} + \int_{0}^{T} \omega'(t) KL\big(q(x_t|y) \| p(x_t)\big) dt  \\
    &= \int_{0}^{T} \omega'(t) \mathbbm{E}_{q(x_t|y)} \Big[ \log \frac{q(x_t| y)}{p(x_t)} \Big] dt 
\end{align*}
where $\omega'(t):= \frac{d\omega(t)} {dt}$. The equality in (a) holds because $\omega(t) KL\big(q(x_t|y) \| p(x_t)\big) \Big]_{0}^{T}$ is zero at $t=0$ and $t=T$. This is because $\omega(t)=0$ by assumption at $t=0$, and $x_T$ becomes a pure Gaussian noise at the end of the diffusion process which makes $p(x_T)=q(x_T|y)$ and thus $KL\big(q(x_T|y) \| p(x_T)\big) = 0$.


For simplicity, let us suppose $\sigma=0$ in the variational distribution, and thus $x_0=\mu$ is deterministic. The forward diffusion process is then $x_t = \alpha_t \mu + \sigma_t \epsilon$ for $\epsilon \sim \mathcal{N}(0,1)$. Applying the re-parameterization trick, the gradient w.r.t $\mu$ can be simply written as 
\begin{align*}
    \nabla_{\mu} {\rm reg}(\mu)  & = 2 \sigma_v^2 \int_{0}^{T} \omega'(t)  \mathbbm{E}_{\epsilon \sim \mathcal{N}(0,1)}  \Big[ \big(\nabla_{x_t} \log q_t(x_t|y) - \nabla_{x_t} \log p(x_t) \big)^{\top} \frac{d x_t}{d \mu} \Big] dt \\
    & = 2 \sigma_v^2 \int_{0}^{T} \omega'(t) \mathbbm{E}_{\epsilon \sim \mathcal{N}(0,1)} \Big[ \big((- \frac{\epsilon}{\sigma_t}) - (- \frac{\epsilon_{\theta}(x_t;t)}{\sigma_t}) \big)^{\top} (\alpha_t I) \Big] dt
\end{align*}
where $q(x_t|y)=\mathcal{N}(\alpha_t \mu, \sigma_t^2 I)$, and $\nabla_{x_t} \log q_t(x_t|y) = - (x_t- \alpha_t \mu)/\sigma_t^2 = -\epsilon / \sigma_t$. Note, the gradient was exchanged with the expectation since both gradient terms exist and are bounded. Finally, we can rearrange terms to arrive at the compact form
\begin{align*}
    \nabla_{\mu} {\rm reg}(\mu) & = \frac{1}{T}\int_{0}^{T} T \omega'(t) \frac{2 \sigma_v^2 \alpha_t}{\sigma_t} \mathbbm{E}_{\epsilon \sim \mathcal{N}(0,1)} \big[ (\epsilon_{\theta}(x_t;t) - \epsilon) \big] dt  \\
    & =  \mathbbm{E}_{t \sim \mathcal{U}[0,T],\epsilon \sim \mathcal{N}(0,1)} \big[ \lambda_t (\epsilon_{\theta}(x_t;t) - \epsilon) \big]
\end{align*}
for $\lambda_t:=T \omega'(t)2 \sigma_v^2\alpha_t/\sigma_t$. Note that we can ignore the second term inside the expectation since $\epsilon$ has a zero mean.


\subsection{Adding dispersion to variational approximation}
The results in Proposition 2 are presented for Dirac distribution with no dispersion, namely $q(x_0|y) \sim \delta(x_0 - \mu)$. We however can easily extend those to optimize for Gaussian dispersion as well. The interesting observation is that the gradient w.r.t. the mean $\mu$ remains the same, and the gradient w.r.t. $\sigma$ is simple and tractable.

In this case, we have $q(x_0|y) \sim \mathcal{N}(\mu, \sigma^2 I)$. The diffusion signal at timestep $t$ can then be represented in a compact form using reparameterization trick as $x_t=\alpha_t \mu+\sqrt{\alpha_t^2 \sigma^2 + \sigma_t^2} \epsilon$. Let's define $\eta_t:=(1 + \sigma^2  (\frac{\alpha_t}{\sigma_t})^2)^{1/2}$ so that $x_t=\alpha_t \mu+ \eta_t \sigma_t \epsilon$. Note, for no dispersion case $\eta_t=1$. Then, gradient w.r.t. the mean is obtained as
\begin{align*}
 \nabla_{\mu} {\rm reg}(\mu, \sigma) &= \int_{0}^{T} \omega'(t)  \mathbbm{E}_{\epsilon \sim \mathcal{N}(0,1)}  \Big[ \big(\nabla_{x_t} \log q_t(x_t|y) - \nabla_{x_t} \log p(x_t) \big)^{\top} \frac{d x_t}{d \mu} \Big] dt \\
& = \int_{0}^{T} \omega'(t) \mathbbm{E}_{\epsilon \sim \mathcal{N}(0,1)} \Big[ \big((- \frac{\epsilon}{\sqrt{\alpha_t \sigma^2 + \sigma_t^2}}) - (- \frac{\epsilon_{\theta}(x_t;t)}{\sigma_t}) \big)^{\top} (\alpha_t I) \Big] dt \\
& = \int_{0}^{T} \omega'(t) \mathbbm{E}_{\epsilon \sim \mathcal{N}(0,1)} \frac{\alpha_t}{\sigma_t} \Big[ \epsilon_{\theta}(x_t;t) - \eta_t^{-1} \epsilon  \Big] dt \\
& = \mathbbm{E}_{\epsilon,t} \big[ \lambda_t \epsilon_{\theta}(x_t;t) \big]
\end{align*}
Similarly, the gradient w.r.t. the dispersion is found as
\begin{align*}
\nabla_{\sigma} {\rm reg}(\mu, & \sigma) = \int_{0}^{T} \omega'(t)  \mathbbm{E}_{\epsilon \sim \mathcal{N}(0,1)}  \Big[ \big(\nabla_{x_t} \log q_t(x_t|y) - \nabla_{x_t} \log p(x_t) \big)^{\top} \frac{d x_t}{d \sigma} \Big] dt \\
& = \int_{0}^{T} \omega'(t) \mathbbm{E}_{\epsilon \sim \mathcal{N}(0,1)} \Big[ \big((- \frac{\epsilon}{\sqrt{\alpha_t \sigma^2 + \sigma_t^2}}) - (- \frac{\epsilon_{\theta}(x_t;t)}{\sigma_t}) \big)^{\top} \epsilon (\alpha_t \sigma^2 + \sigma_t^2)^{-1/2} 2 \alpha_t^2 \sigma \Big] dt \\
& = \sigma \mathbbm{E}_{\epsilon,t} \Big[ \lambda_t 2 \eta_t^{-1} \big(\frac{ \alpha_t }{\sigma_t} \big)    \epsilon^{\top} (\epsilon_{\theta}(x_t;t) - \eta_t^{-1} \epsilon) \Big] 
\end{align*}
The dispersion gradient has close form. However, we believe the Gaussian dispersion is not the right choice to add stochasticity and diversity to natural images since simply perturbing an image with Gaussian noise does not lead to another legitimate image. In essence, one needs a proper dispersion on the image manifold that needs more sophisticated dispersion models such as in \cite{wang2023prolificdreamer}.

\section{Additional experiments}

\subsection{Pretrained diffusion model}
We adopt the score function from the pretrained guided diffusion model, which uses no class conditioning. It is trained on $256 \times 256$ ImageNet dataset. Architecture details are listed in section 3 of \cite{dhariwal2021diffusion}.

\subsection{Image superresolution}
\begin{figure}
    \centering
        \hspace{0mm}\includegraphics[width=1.0\textwidth]{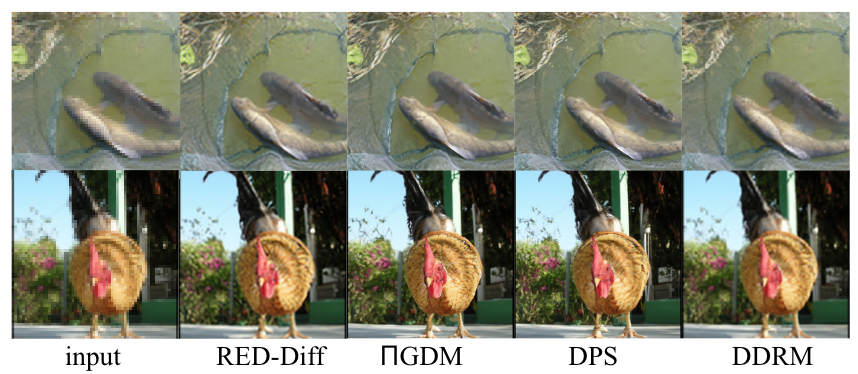}  
\caption{Comparison of the proposed variational sampler with alternatives for superresolution of representative ImageNet examples. Each sampler is tuned for the best performance. }  \label{fig:sr_comp}
\end{figure}

We perform 4x superresolution on a 1k subset of ImageNet dataset. Bicubic degradation is applied to create the low-resolution inputs. After tuning $\lambda=0.25$, Adam iterations are run for $1,000$ steps. A few samples are illustrated in Fig.~\ref{fig:sr_comp}, where one can notice that RED-diff strikes a good balance between image fidelity and perceptual quality. One can play with this trade-off by tuning the Adam learning rate. Smaller learning rates lead to higher fidelity, while larger learning rates give rise to higher perceptual quality. See supplementary material for more examples.



\begin{table}
    \centering
    \renewcommand\tabcolsep{6pt}
    \caption{\small Performance of different samplers for ImageNet 4x superresolution. We adopt unconditional guided diffusion model for the score function. We choose Adam $lr=0.25$. }
    
    \vspace{4mm}
    \scalebox{1.0}{
    \centering
    \begin{tabular}{l|ccccc}
    Sampler & PSNR(dB) $\uparrow$ & SSIM $\uparrow$ & KID $\downarrow$ & LPIPS $\downarrow$  & top-1 $\uparrow$  \\
    \hline
    \hline
    \rule{0pt}{2ex} 
    DPS & 24.83 & 0.71 &  10.01 & 0.16 & \bf{71.5}    \\
    $\Pi$GDM & 25.25 & 0.73 &  10.9 & \bf{0.15} & 71.02   \\
    DDRM & 25.32 & 0.72 &  14.0 & 0.23 & 63.9    \\
    RED-diff & \bf{25.95} & \bf{0.75} & \bf{10.0} & 0.25 & 66.7   \\
    \hline
    \end{tabular}
    }
    \label{tab:eval_sr}
\end{table}

\begin{figure}[t]
\begin{center}
   \includegraphics[width=0.75\linewidth]{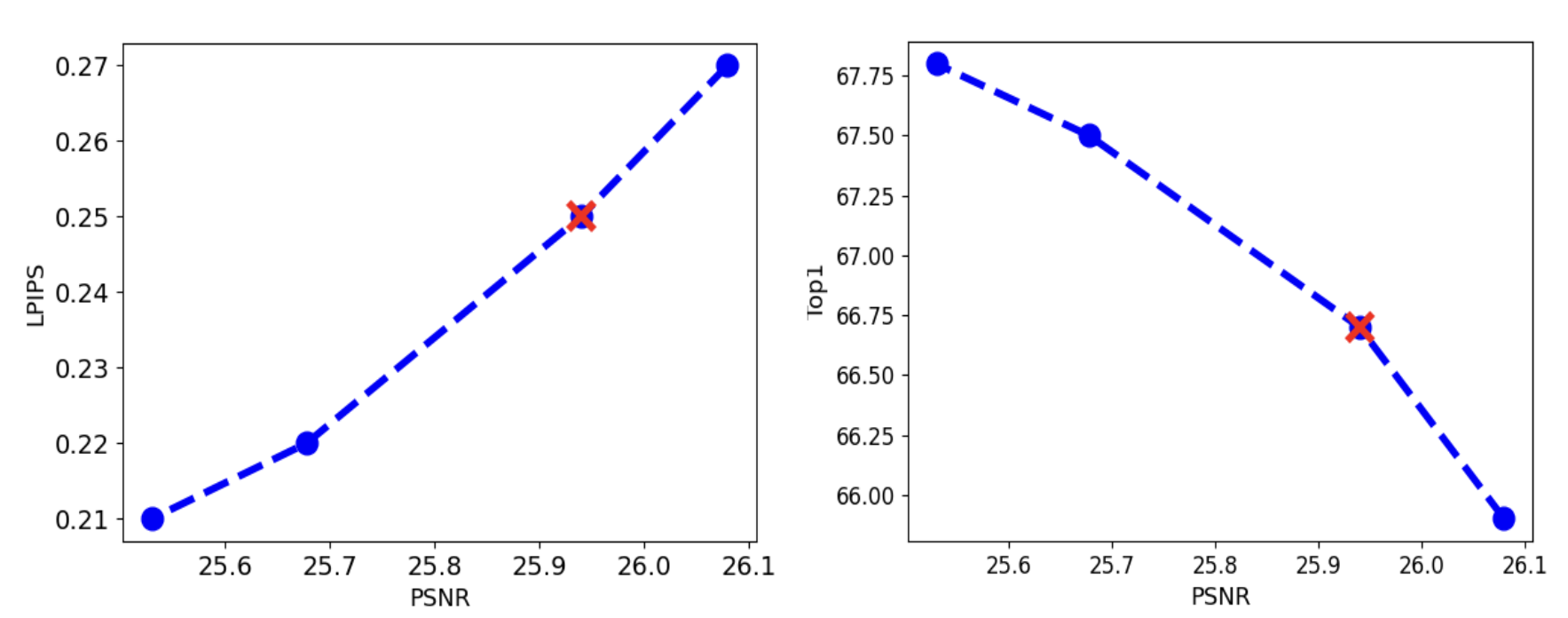}
\end{center}
   \caption{LPIPS vs. PSNR (dB) trade-off for 4x superresolution.}
\label{fig:psnr_lpips}
\end{figure}

Quantitative results are also listed in Table \ref{tab:eval_sr}. One can see that RED-diff significantly outperforms the alternative samplers in terms of PSNR and SSIM. Table \ref{tab:eval_sr} however indicates that RED-diff is not as good as other alternatives in terms of the perceptual quality. We want to note the trade off we observe between fidelity (e.g., PSNR) and perceptual quality (e.g., LPIPS). It seems that one can achieve better perceptual quality at the expense of lower fidelity by tuning the regurlization weight $\lambda$ that control the bias-variance trade off. Other hyperparamaters such as step size and number of steps can be tuned alternatively. Fig. \ref{fig:psnr_lpips} depicts the trade off between PSNR-LPIPS, where the red-cross denotes the point reported in Table \ref{tab:eval_sr}.

\subsection{More examples for comparisons}
\begin{figure*}
    \centering
        \includegraphics[width=1.0\textwidth]{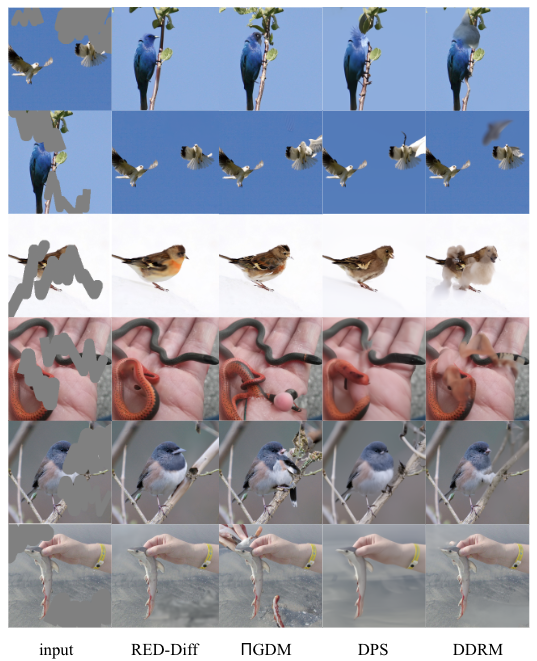}  
\caption{Comparison of the proposed variational sampler with alternatives for inpainting representative ImageNet examples. Each sampler is tuned for the best performance. }  \label{fig:inp_comp_more}
\end{figure*}

\begin{figure*}
    \centering
        \includegraphics[width=1.0\textwidth]{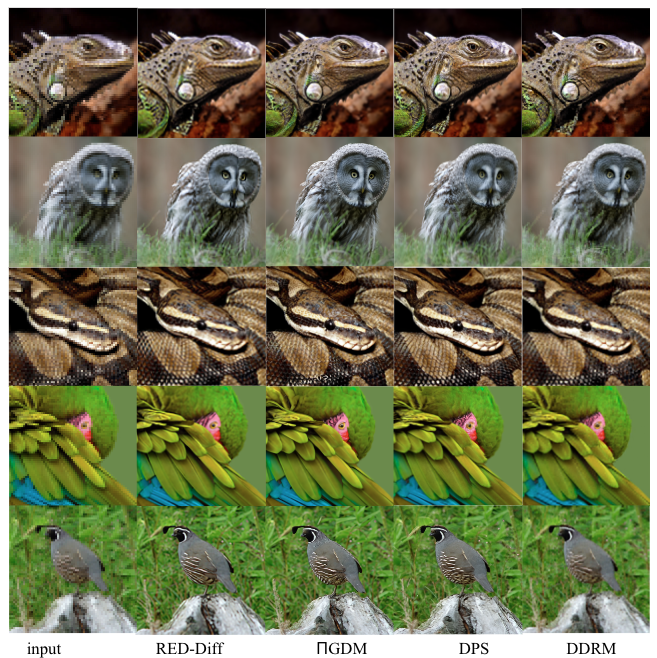}  
\caption{Comparison of the proposed variational sampler with alternatives for superresolution of representative ImageNet examples. Each sampler is tuned for the best performance. }  \label{fig:sr_comp_apdx}
\end{figure*}

We provide additional examples to compare RED-diff with alternative methods for inpainting and superresolution. The inpainting examples are shown in Fig. \ref{fig:inp_comp_more}. Superresolution examples are also shown in Fig. \ref{fig:sr_comp_apdx}. For both tasks, the same setup as discussed n the main paper was used for each scheme.

 \subsection{Diversity}
\begin{figure*}
    \centering
        \includegraphics[width=1.0\textwidth]{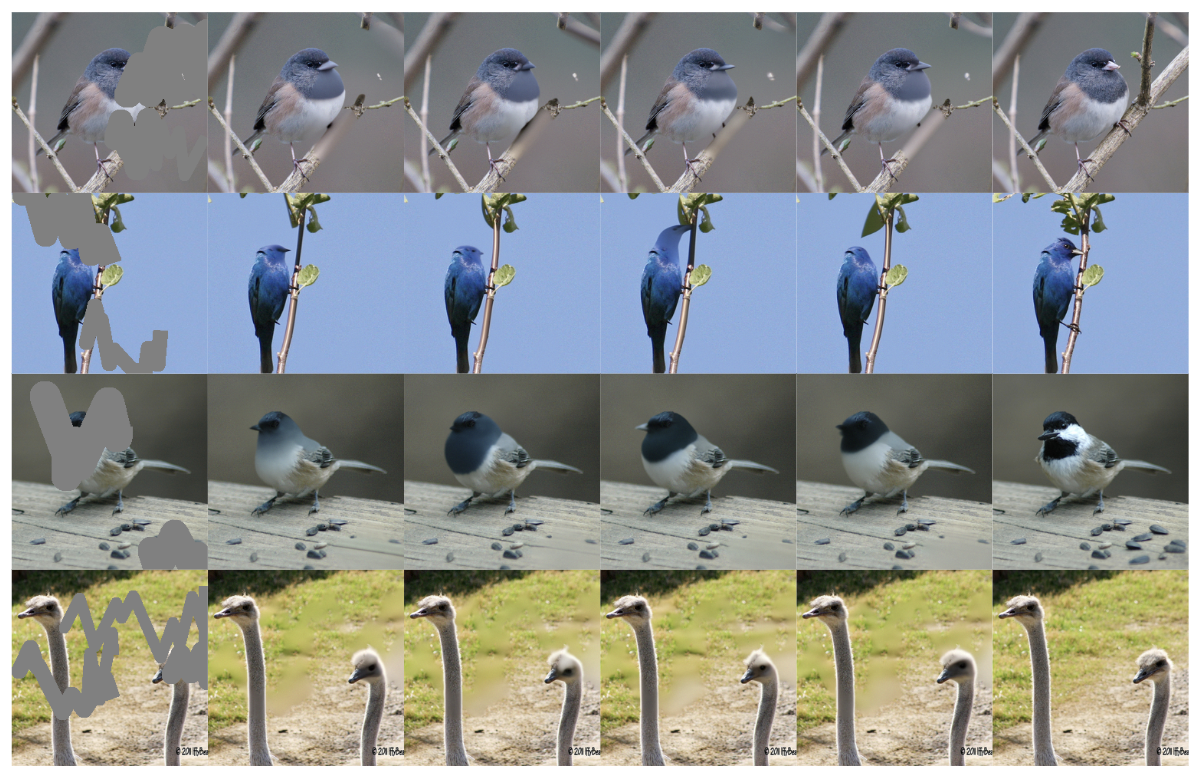}   
\caption{Sampling diversity for ImageNet inpainting. A few random realizations are shown, where the first column is the masked input image, and the last one is the reference image.  }  \label{fig:div_inp}
\end{figure*}

To verify the sample diversity for RED-diff, Fig. \ref{fig:div_inp} illustrates different samples for the ImageNet inpainting task when the seed for $\epsilon$ is changing. We choose Adam optimizer with $lr=0.25$, and $1,000$ steps. From the examples, we observe that RED-diff samples are sufficiently diverse. To further enhance the diversity, one can play with the optimizer parameters, for example by choosing a larger learning rate for Adam, or using a smaller number of steps.

 \subsection{Diffusion evolution}
\begin{figure*}
    \centering
        \includegraphics[width=1.0\textwidth]{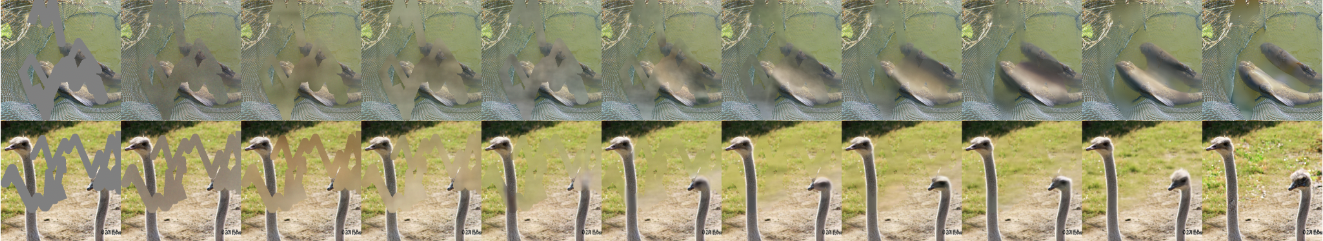}   
\caption{Evolution of RED-diff over iterations. Descending sampling direction is used. Denoisers from large to small $t$ restore high-level to low-level features, respectively. Adam with $1,000$ steps used and every $100$ iterations are visualized. }  \label{fig:evol_diff}
\end{figure*}

\begin{figure}
    \centering
        \includegraphics[width=1\textwidth]{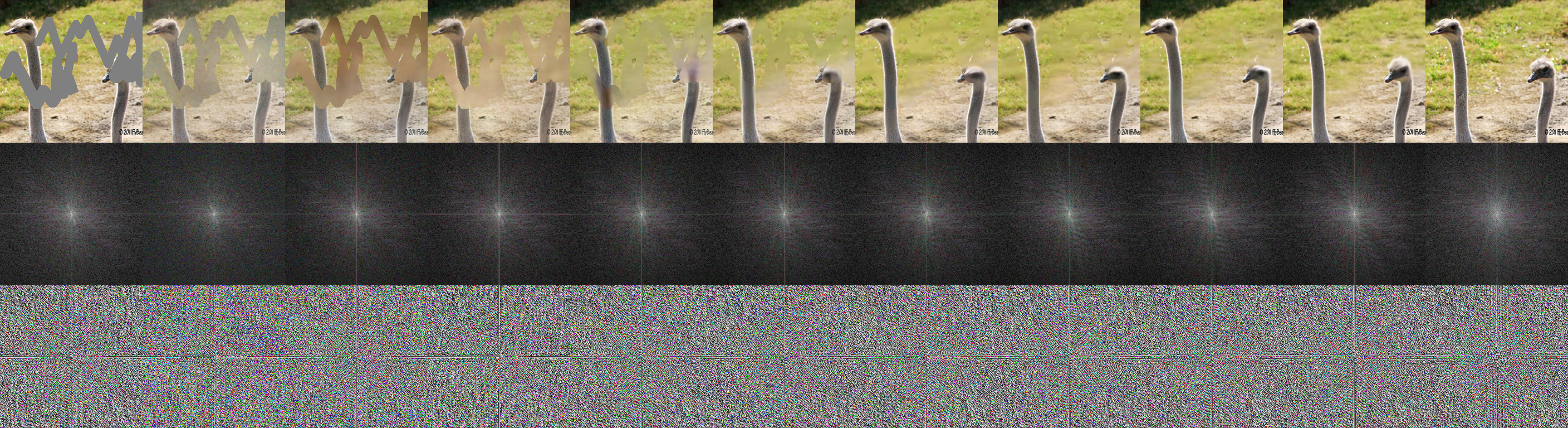}   
\caption{Evolution of RED-diff over iterations in the frequency domain. Middle row shows the log-magnitude, and the bottom row shows the phase.  }  \label{fig:fft_evol}
\end{figure}

To understand the restoration process with diffusions, we plot the evolution of the diffusion model over timesteps in Fig. \ref{fig:evol_diff}. We visualize $\hat{\mu}$ as the outputs from every $100$ steps over $1,000$ timesteps. To gain further insights into the generation process and how the image structures are generated, the evolution of diffusion steps is also plotted in the frequency domain. Fig. \ref{fig:fft_evol} illustrates the magnitude and phase every $100$ steps over $1,000$ timesteps. It can be observed that the earlier denoisers add high frequency details, while the later denoisers generate low frequency structures.

\section{Additional scenarios}
We include additional tasks to assess RED-diff for noisy inverse problems. We consider noisy inpainting and compressed sensing MRI.

\subsection{Compressed sensing MRI}
\begin{figure*}[t]
\begin{center}
   \includegraphics[width=1.0\linewidth]{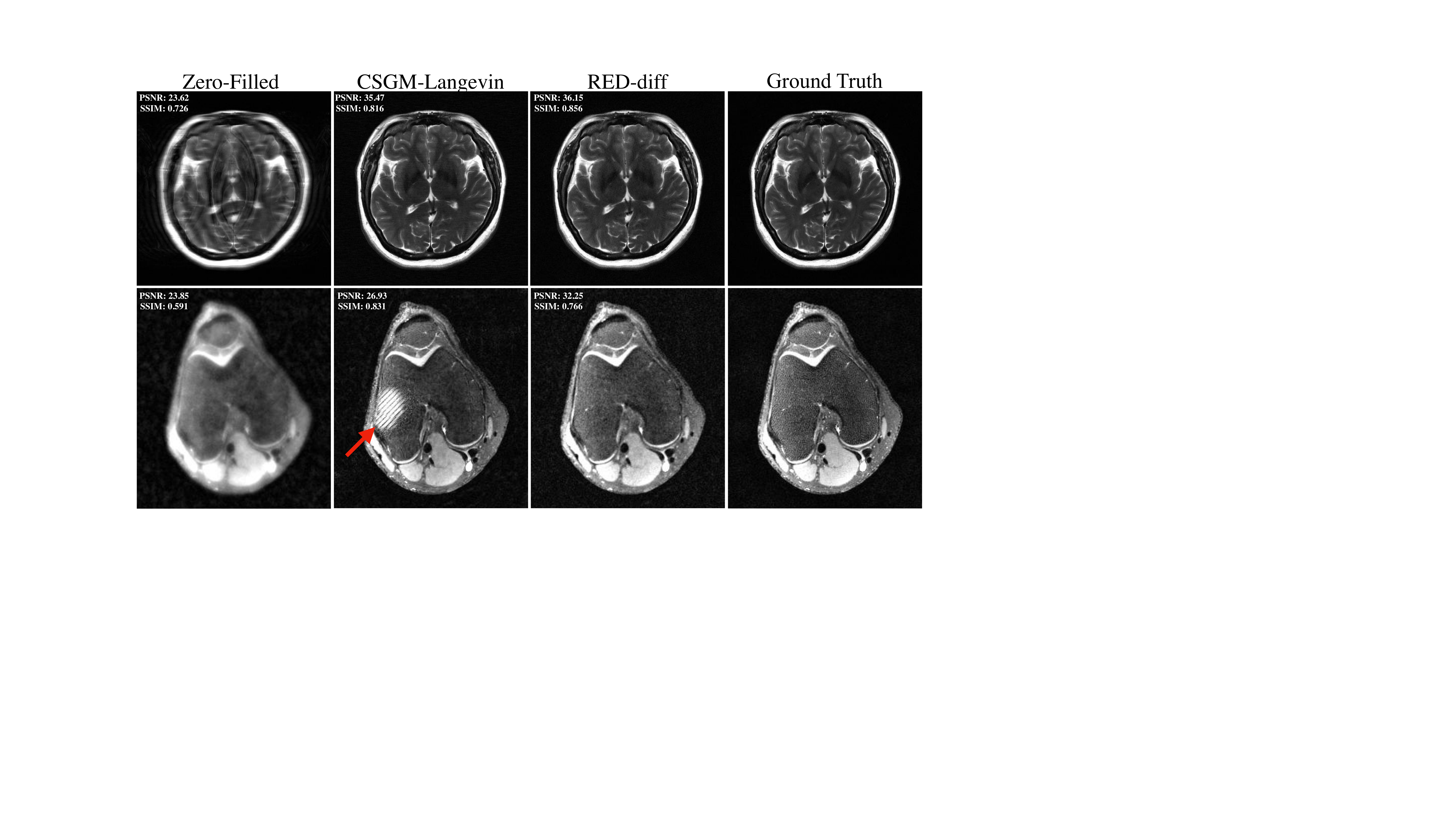}
\end{center}
   \caption{Representative reconstructed images for brain with $R=4$, and knee with $R=12$.}
\label{fig:fig_cs_mri}
\end{figure*}

We adopt the pretrained diffusion model from \cite{jalal2021robust}, and sample via RED-Diff. The pretrained diffusion has been trained based on fastMRI brain data. We test sampling for both in-domain brain, and out-of-domain knee data. The performance is compared with CSGM-Langevin \cite{jalal2021robust} for which the codebase is publicly available, and serves as a state-of-the-art for complex-valued medical image reconstruction. We use the multi-coil fastMRI brain dataset \cite{zbontar2018fastmri} with 1D equispaced undersampling, and the fully-sampled 3D fast-spin echo multi-coil knee MRI dataset from \cite{ong2018mridata} with 2D Poisson Disc undersampling mask, as in \cite{jalal2021robust}. We used $6$ validation volumes for fastMRI, and $3$ volumes for Mridata by selecting $32$ middle slices from each volume. We use exactly the same tuning parameters as for the inpainting task. The results for different undersampling rates (R) are shown in Table \ref{tab:cs_mri}. Obviously, RED-Diff outperforms CSGM method (with Langevin sampling) consistently in terms of PSNR, that is the measure of interest in MRI reconstruction. For more details see \cite{ozturkler2023regularization}. Representative samples are also shown in Fig. \ref{fig:fig_cs_mri}.

\begin{table}[t!]
\centering
\renewcommand\tabcolsep{6pt}
\caption{Compressed sensing MRI PSNR (dB) for fastMRI brain and Mridata knee dataset with Rx undersampling. CSGM is a SOTA method for complex-valued MRI with a public codebase. }
\vspace{3mm}
\scalebox{0.99}{
\centering
\begin{tabular}{l|c|cc|c}
Anatomy &  \multicolumn{1}{c|}{Brain} & \multicolumn{2}{c|}{Knee} & {time} \\
\hline\hline     \rule{0pt}{2ex} 
Sampler &  $R = 4$ &  $R = 12$ & $R = 16$ & (sec/iter) \\
\hline \rule{0pt}{2ex}
CSGM & 36.3 &  31.4 & 31.8 & 0.344\\ 
RED-diff  & \textbf{37.1} &  \textbf{33.2} & \textbf{32.7} & \textbf{0.114}\\
\hline
\end{tabular}
}
\label{tab:cs_mri}
\end{table}

\subsection{Noisy inpainting}
To see the effects of measurement noise on RED-diff performance, we add Gaussian noise with $\sigma_v=0.1$ to the masked Palette images for inpainting. We compare RED-diff with DPS and PGDM, where for all we use $100$ steps. For RED-diff we choose $\lambda=0.25$ similar to all other scenarios with $lr=0.25$. For DPS, we choose $\eta=0.5$, and the step size run $\zeta = 0.5/\|y - A(\hat{x}_0(x))\|$ adopted from \cite{chung2022diffusion} where we run a grid search over the range $[0,1]$ for the coeffficient. Table \ref{tab:eval_inp_noisy} shows the metrics. It is seen that RED-diff outperforms DPS and $Pi$GDM in most metrics.

\begin{table*}
    \centering
    \renewcommand\tabcolsep{6pt}
    \caption{\footnotesize Noisy inpainting with $\sigma_y=0.1$ for Palette data with 100 steps.}
    \vspace{3mm}
    \scalebox{0.99}{
    \centering
    \begin{tabular}{l|ccccc}
    Sampler & PSNR(dB) $\uparrow$ & SSIM $\uparrow$ & KID $\downarrow$ & LPIPS $\downarrow$  & top-1 $\uparrow$  \\
    \hline
    \hline
    \rule{0pt}{2ex} 
    DPS & 16.98 & 0.39 &  \textbf{1.9} & 0.52 & 8.8    \\
    $\Pi$GDM & 16.15 & 0.31 &  3.6 & 0.45 & 42.4    \\
    RED-diff & \bf{18.92} & \bf{0.41} & 3.4 & \bf{0.4} & \bf{43.3}   \\
    \hline
    \end{tabular}
    }
    \label{tab:eval_inp_noisy}
\end{table*}

 \section{Ablations}

\subsection{Denoiser weighting mechanism}
\begin{figure*}
    \centering
    \begin{subfigure} 
        \centering
        \includegraphics[width=0.7\textwidth]{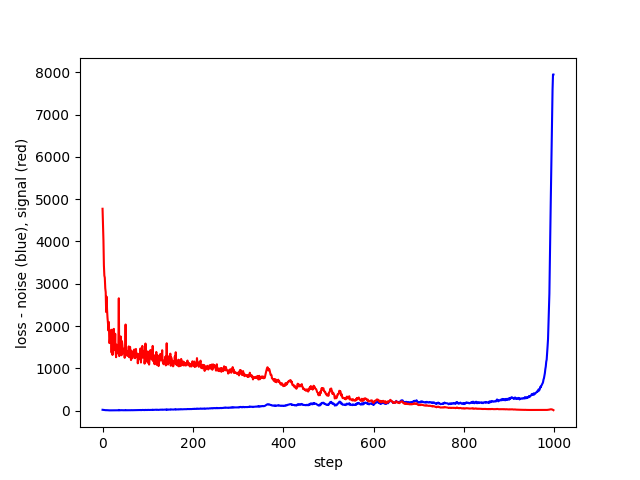}
    \end{subfigure}
\caption{Evolution of signal residual $\|\hat{\mu}_t - \mu\|^2$ (red) and noise residual $\|\epsilon_{\theta}(x_t;t) - \epsilon\|^2$ (blue) over time for ImageNet inpainting when equally weighting diffusion denoisers at different timesteps.  }\label{fig:loss}
\end{figure*}

It was discussed in section 4.3 that the noise residual blows up at the last timesteps of the diffusion process. Here, we illustrate that in Fig. \ref{fig:loss} when denoisers are equally weighted. It shows the magnitude of both the signal and noise residuals. It is clearly seen that the noise residual goes up drastically at around $t=1,000$. Similarly, the signal residual also blows up at earlier steps. This suggests a mechanism that guarantees a relatively small signal residual at all iterations that led to the SNR ruler for weighting the denoisers.

 \subsection{Optimizing RED-diff for more epochs}
\begin{figure*}
    \centering
        \includegraphics[width=1.0\textwidth]{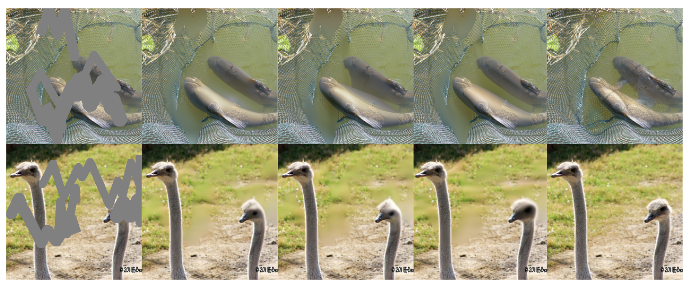}   
\caption{RED-diff outputs for a different number of epochs. From left to right: input, epochs=1,2,10, and the reference, respectively. Adding more epochs does not make any noticeable difference.  }  \label{fig:inp_epochs}
\end{figure*}

Since RED-diff is an optimization-based sampling, one may ponder that using more iterations by going over the denoisers more than once can improve the sample quality. To test this idea, we choose different epochs for ImageNet inpainting, when we use Adam optimizer with $1,000$ steps per epoch. A few representative examples are shown in Fig. \ref{fig:inp_epochs}. It is observed that adding more epochs has negligible improvement on the sample quality.

\subsection{Optimization strategy}
\begin{figure}
    \centering
        \includegraphics[width=1.0\textwidth]{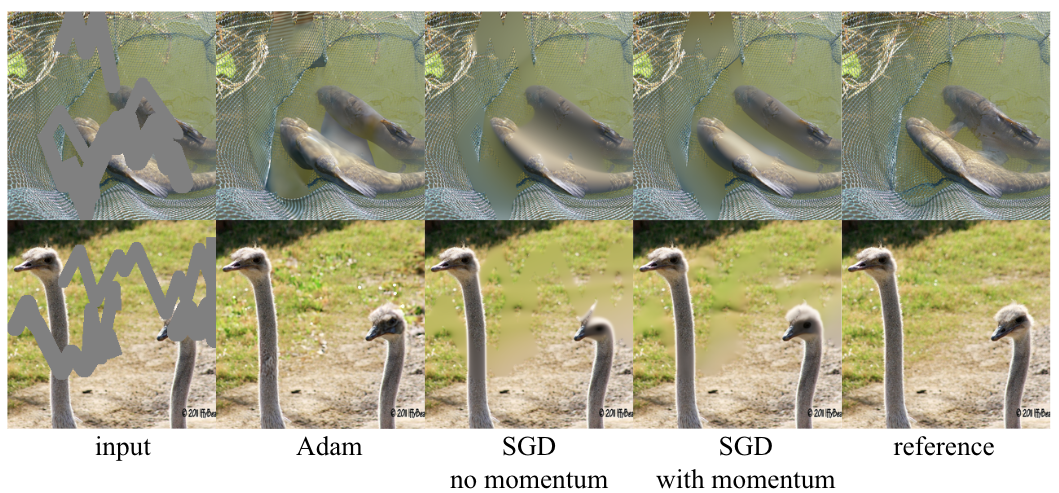} 
\caption{Ablation of optimization strategy for an inpainting example from ImageNet dataset. Adam and SGD (with momentum $0.9$) seem to perform similarly.   }  \label{fig:ablate_opt}
\end{figure}

The proposed variational sampler relies on optimization for sampling. To see the role of optimizer, we first ablate SGD versus Adam. We found SGD more sensitive to step size which in turn demands more careful tuning. We tune the hyperparameters in each case for the best performance for ImageNet inpainting with $\lambda=0.25$. We use Adam with parameters discussed in the beginning of Section~\ref{sec:exps}. Fig.~\ref{fig:ablate_opt} shows representative inpainting examples, where SGD is tested with momentum ($0.9$) and without momentum. It appears that SGD with momentum can be as good as the Adam optimizer, which indicates that RED-diff is not sensitive to the choice of optimizer. Our future direction will explore accelerated optimization for faster convergence. 

\begin{figure}
    \centering
        \includegraphics[width=1.0\textwidth]{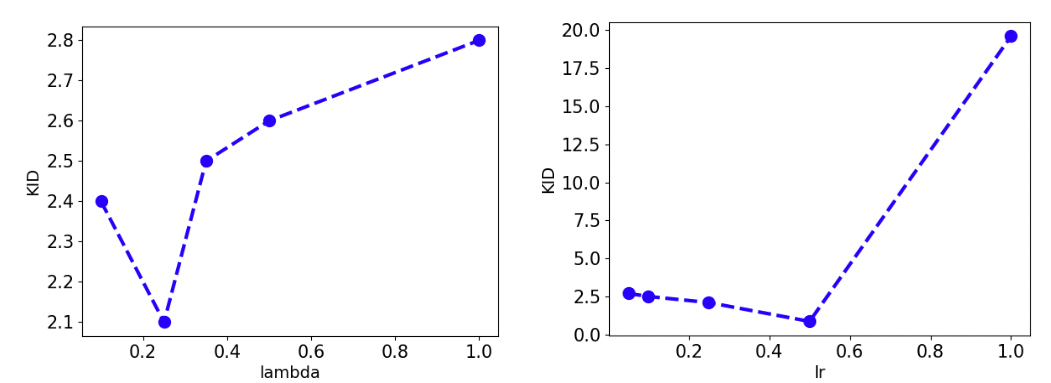}   
\caption{KID versus the optimizer learning rate (right), and the weight $\lambda$ (left).}  \label{fig:kid_vs_lr_lambda}
\end{figure}

For the Adam optimizer, we also ablated the initial learning rate to see its effect on the quality of the generated samples. The results are depicted in Fig.~\ref{fig:kid_vs_lr_lambda} (right). It appears that tuning Adam learning rate is important, where a smaller learning rate (e.g., $0.05$) leads to better reconstruction quality since the optimizer can better converge to the (optimal) MAP estimate. However, using a larger learning rate (e.g., $0.5$) leads to better perceptual quality measured with KID.

KID for a range of $\lambda$ is also depicted in Fig.~\ref{fig:kid_vs_lr_lambda} (left). It is evident that there is an optimal $\lambda$, which confirms that with proper tuning of the bias-variance trade-off one can gauge the sampling quality.


\subsection{Timestep sampling strategy} 
\label{subsec: ablate_sampling}
\begin{figure*}
    \centering
        \includegraphics[width=1.0\textwidth]{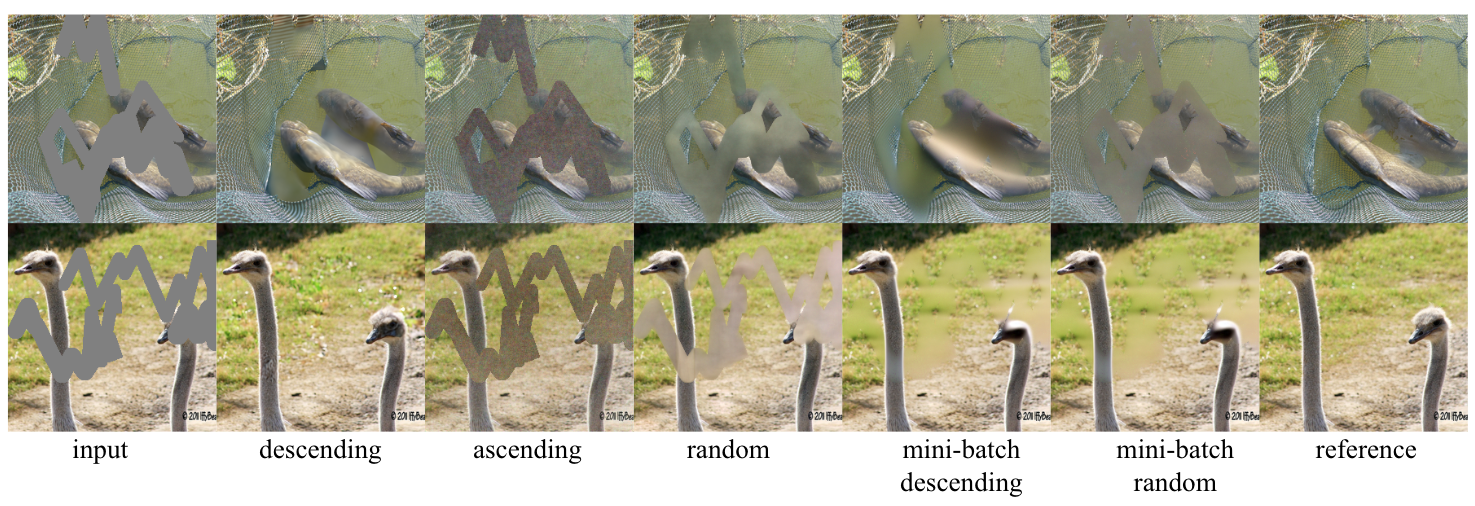}   
\caption{\small Ablation of sampling strategy for ImageNet inpainting. For mini-batch sampling, the batch size is 25. Sampling in descending direction performs significantly better than the rest. Adam optimizer is used with an unconditional ImageNet guided diffusion score. } \label{fig:ablate_sampling}
\end{figure*}

We ablate the timestep sampling strategy and the number of sampling steps. For comparison, we consider five different strategies when sampling from $t$ during optimization. This includes (1) random sampling; (2) ascending; (3) descending; (4) min-batch random sampling; and (5) mini-batch descending sampling. We adopt Adam optimizer with $1,000$ steps and choose the linear weighting mechanism with $\lambda=0.25$. Random sampling (1) uniformly selects a timestep $t \in [1,T]$, while ascending and descending sampling are ordered over timesteps. As shown in the inpainting example in Fig.~\ref{fig:ablate_sampling}, it is seen that descending sampling performs significantly better than others. It starts from the denoiser at time $t=T$, adding semantic structures initially, and then fine details are gradually added in the process. This appears to generate images with high fidelity and perceptual quality.

The aforementioned sampling strategies choose a single denoiser at a time, but one may ponder what if one performs batch sampling to regularize based on multiple denoisers at the same time. This is also computationally appealing as it can be executed in parallel based on our optimization framework. To test this idea, we sort $1,000$ time steps in descending order, and use a batch of $25$ denoisers per iteration, for $40$ iterations. It is again observed from Fig.~\ref{fig:ablate_sampling} that batch sampling smears the fine texture details. This needs further investigation to find out the proper weighting mechanism that enables parallel sampling. In conclusion, descending sampling with a single (or a few) denoiser at a time seems to be the best sampling strategy. More details are found in the supplementary material.

\begin{table}[ht]
\centering
    \renewcommand\tabcolsep{5pt}
    \caption{\small Performance of RED-diff for inpainting under different step counts, when $\lambda=0.25$ and $lr=0.25$.}
    \vspace{3mm}
    \scalebox{1.0}{
    \begin{tabular}{l|ccccc}
    steps & PSNR(dB) $\uparrow$ & SSIM $\uparrow$ & KID $\downarrow$ & LPIPS $\downarrow$  & top-1 $\uparrow$  \\
    \hline
    \hline
    \rule{0pt}{2ex} 
    10 & 18.83 & 0.75 &  22.05 & 0.21 & 59.4    \\
    100 & 23.13 & 0.87 &  \bf{1.93} & 0.12 & \bf{70.8}    \\
    1000 & 23.86 & 0.88 & 2.17 & 0.11 & 69.8  \\
    \hline
    \end{tabular}
    }
    \label{tab:eval_steps}
\end{table}

It is also useful to understand how many steps RED-diff needs to generate good samples. To this end, we evaluated ImageNet inpainting for a different number of steps in Table~\ref{tab:eval_steps}. One can observe that with $100$ steps the best perceptual quality is achieved where KID=1.93. With more steps, the optimizer tends to refine the fidelity so a better reconstruction PSNR is achieved. This ablation suggests that not many steps are needed for the variational sampler. This is in contrast with DreamFusion \cite{poole2022dreamfusion} which uses a large number of iterations (15k) to generate samples.


    

\subsection{Connection and differences with RED}
\begin{figure*}[t]
\begin{center}
   \includegraphics[width=1.0\linewidth]{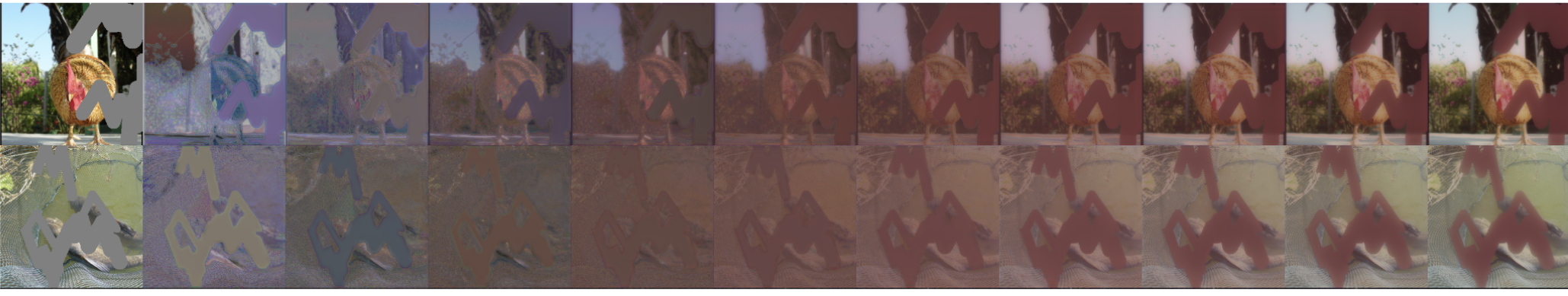}
\end{center}
   \caption{Evolution of diffusion when the noise to each denoiser is removed and $x_t=\mu$.}
\label{fig:evol_reddiff_no_noise}
\end{figure*}

We want to further elaborate the connections with the RED framework. It is useful to recognize that the classical RED adds no noise to the input for denoising, and it is simply a fixed-point problem. RED also uses a single {\it deterministic} denoiser. RED-diff is however fundamentally different. It is generative and add noise to the input of all denoisers in the diffusion trajectory. As a result it stochastically navigates towards the prior. This seems crucial to find a plausible solution. To highlight of the strength of the RED-diff regulrization over RED we performed two simple experiments. In the first experiment, we removed noise from the input of all denosiers, namely $x_t=\mu_t$, and then ran RED-diff iterations for the image inpainting task. It appears that iterations do not progress and they get stuck around the initial masked image as seen in Fig. ~\ref{fig:evol_reddiff_no_noise}. In the second experiment, we just used a single denoiser at time $t=0$, which has a very small noise, and thus resembles RED the most. We observe again that the solution gets such and cannot navigate to the real region of prior to fill out the masked areas in the image. We tested for a single denoiser picked from other time steps such as $t=500$, and observed the same behavior.

\end{document}